\documentclass{article}
\usepackage{spconf,amsmath,graphicx}

\title{DMSA: DYNAMIC MULtI-SCALE UNSUPERVISED SEMANTIC SEGMENTATION BASED ON ADAPTIVE AFFINITY}
 \name{Kun Yang, Jun Lu\sthanks{Corresponding Author: lujun111\_lily@sina.com.}}
 
\address{College of Computer Science and Technology, Heilongjiang University, 150080 Harbin, China \\Jiaxiang Industrial Technology Research Institute of HLJU, Jining, Shandong Province, China\\
Key Laboratory of Database and Parallel Computing of Heilongjiang Province, Harbin, China
}

\usepackage{bbding}
\usepackage{multirow}
\begin{document}
%
\maketitle
\begin{abstract}
The proposed method in this paper proposes an end-to-end unsupervised semantic segmentation architecture DMSA based on four loss functions. The framework uses Atrous Spatial Pyramid Pooling (ASPP) module to enhance feature extraction. At the same time, a dynamic dilation strategy is designed to better capture multi-scale context information. Secondly, a Pixel-Adaptive Refinement (PAR) module is introduced, which can adaptively refine the initial pseudo labels after feature fusion to obtain high quality pseudo labels. Experiments show that the proposed DSMA framework is superior to the existing methods on the saliency dataset. On the COCO 80 dataset, the MIoU is improved by 2.0, and the accuracy is improved by 5.39. On the Pascal VOC 2012 Augmented dataset, the MIoU is improved by 4.9, and the accuracy is improved by 3.4. In addition, the convergence speed of the model is also greatly improved after the introduction of the PAR module.
\end{abstract}
\section{Introduction}
\label{sec:intro}

Self-supervision can perform representation learning without manual labeling, which is completely unsupervised semantic segmentation. DeepLab v3+ \cite{deeplab} uses ASPP structure and Encoder-Decoder \cite{encodedr-decoder} model. ASPP uses dilated convolution with different dilation rates to improve the receptive field of the network. TransFGU \cite{TransFGU} designed a top-down segmentation model using the Vision Transformer (ViT) \cite{vit} model trained by DINO \cite{dino} and showed excellent results on multiple datasets. The end-to-end weakly supervised semantic segmentation model \cite{affinity} designed a PAR module, which integrated local pixel and position information and was used to extract local RGB information to refine the pseudo labels.

This paper introduces ASPP into ViT to enhance feature extraction. At the same time, a dynamic and variable dilation rate strategy is designed independently to better capture multi-scale feature information. PAR module is introduced after the output of teacher network to strengthen the semantic consistency between adjacent pixels, which is used to refine the quality of pseudo labels and better guide the convergence of the model. In this paper, an end-to-end unsupervised semantic segmentation architecture DMSA based on four loss functions is designed. DMSA achieves good results on the saliency datasets such as Pascal VOC and COCO 80.

\section{Related Work}
\label{sec:format}

Unsupervised semantic segmentation is to obtain pixel-level semantic concepts without manual annotation. IIC \cite{iic} extends the clustering of mutual information to pixel-level representation by outputting the probability map on pixels. PiCIE \cite{picie} introduced photometric invariance and geometric equivariance as inductive biases. This approach is that it only works on dataset MS COCO \cite{mscoco}, which does not distinguish between foreground and background classes. MaskContrast \cite{maskcontrast} first uses the object mask to determine a priori, and then uses the priori obtained from the contrast loss to generate pixel-level embedding and this method can only be applied on salient datasets. 
Some multi-stage methods \cite{multi-stage1,multi-stage2} focus on using Class Activation Maps (CAM) \cite{cam} to obtain the initial pixel-level pseudo labels, such as TransFGU, which first uses the sliding window and the trained ViT model by DINO to process the input image, and uses the obtained CAM as the initial pseudo label. Then the teacher-student network \cite{teacher-student} is used to refine and a good effect is obtained, but it may lose features and cause the segmentation effect to be lower than the true value. In this paper, it studies the inaccurate phenomenon of segmentation caused by the failure to capture the global context information and the loss of edge information in unsupervised semantic segmentation.

\section{Method}
\label{sec:pagestyle}

  \begin{figure}[ht]
\begin{minipage}[b]{1.0\linewidth}
  \centering
  \centerline{\includegraphics[width=8cm]{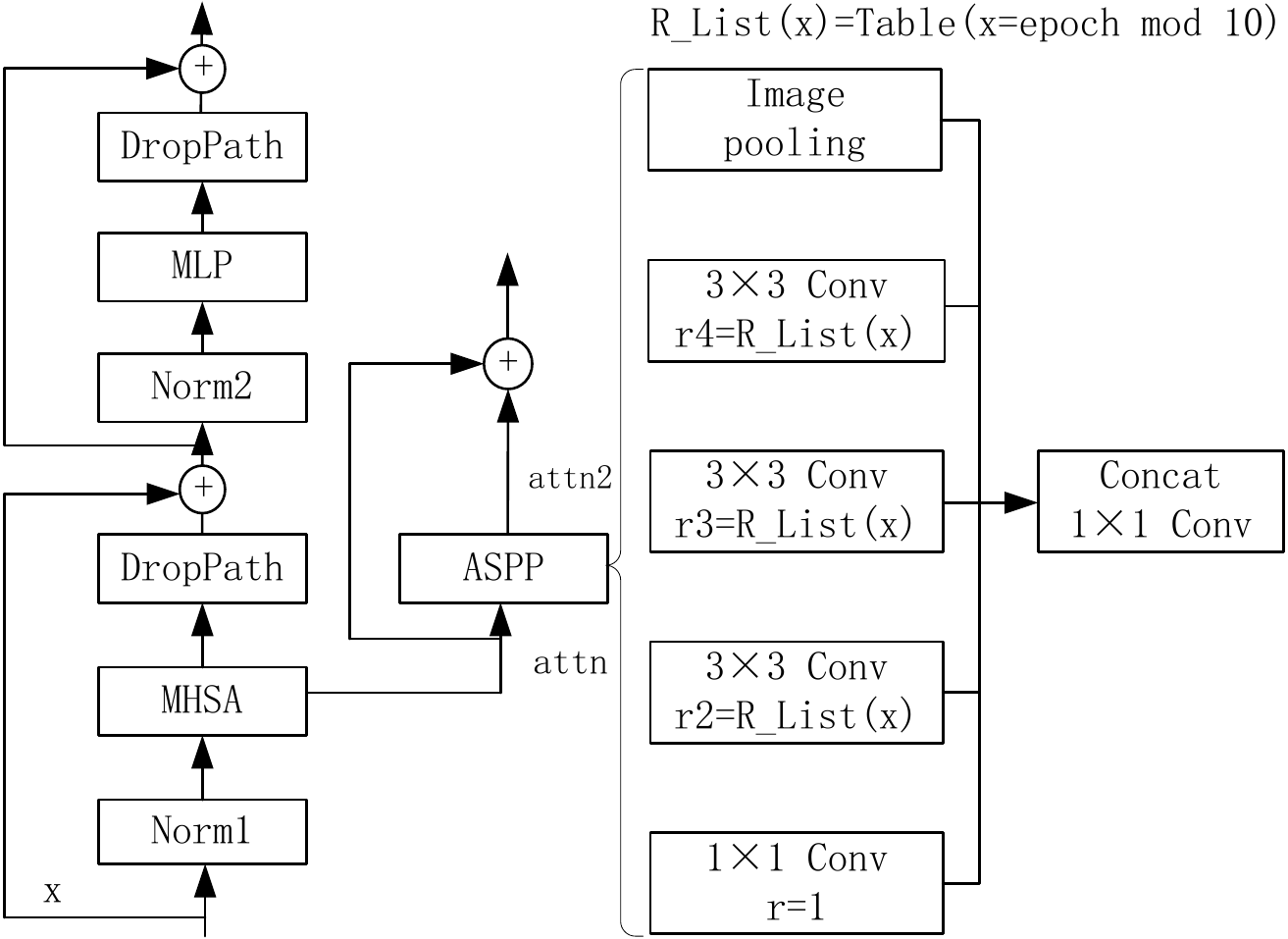}}
\end{minipage}
\caption{The multi-head self-attention output part of Transformer encoder uses ASPP with dynamic dilated rate for multi-scale feature extraction. The R\_List(x) function is a dynamic dilated rate return function, x is the number of epochs, and Table is the dilated rate table.}
\label{fig:res}
\end{figure}
In Figure 1, the Transformer encoder uses the ASPP structure to enhance feature extraction for the attention map output by multi-head self-attention to capture the semantic information on multiple scales. 
\subsection{Dynamic Depth-separable Convolution}
\label{ssec:subhead}

This paper introduces the ASPP structure into the model and designs the original static dilated rate into a variable dynamic dilated rate. The feature map output by ASPP module uses convolution kernel $k'$ of dilated convolution to calculate the feature map, as shown in Equation (1).
\begin{equation}M = \left\lfloor {\frac{{m + 2p - k'}}{s}} \right\rfloor  + 1\end{equation}

In Equation (3), $m$ is the input dilated convolution size, $s$ is the stride, and the obtained dilated convolution feature map size is $M$.   

 \begin{figure*}[ht]
\begin{minipage}[b]{1.0\linewidth}
  \centering
  \centerline{\includegraphics[width=14cm]{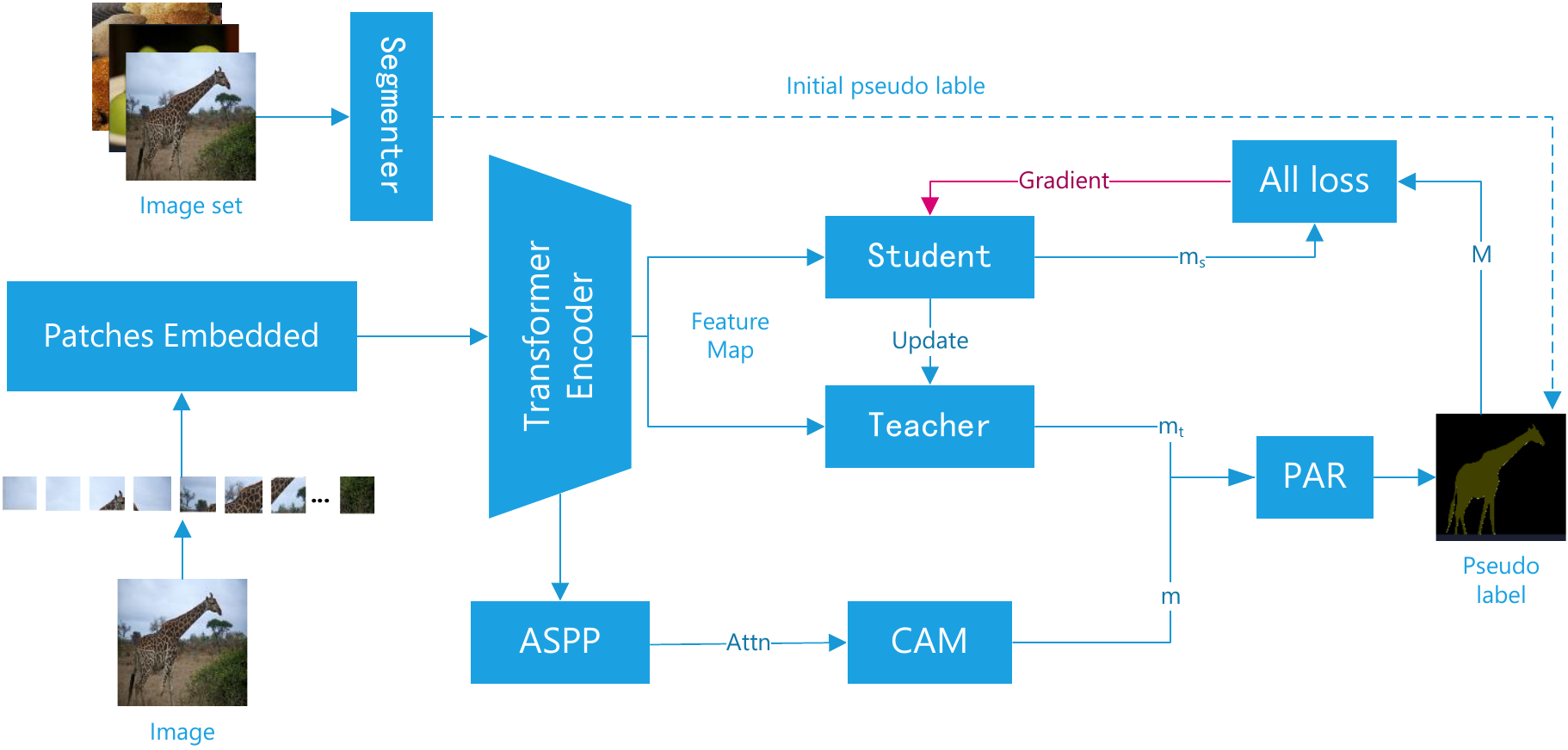}}
\end{minipage}
\caption{DMSA, an end-to-end unsupervised semantic segmentation model. First, CAM is used to get the initial label. The attention map is obtained by Transformer encoder and the feature extraction is strengthened by ASPP. The mask obtained by CAM is fused with the mask obtained by the teacher network. Then the result is refined further by the PAR module to get the final pseudo label. Finally, the mask obtained from the student network and the pseudo label obtained from PAR are used to calculate the loss and gradually improve the training quality.}
\label{fig:res}
\end{figure*}

\subsection{Affinity Refinement in Encoder-Decoder Model}
\label{ssec:subhead}

This paper proposes model DMSA, which uses ViT as the encoder. DMSA only needs the attention map returned by the last attention block, as shown in Equation (2).
 \begin{equation}
attn(Q,K,V) = softmax(\frac{{Q{K^T}}}{{\sqrt {{d_k}} }})V\end{equation}
In Equation (2), $Q$ represents query, which is used to match keys, $K$ represents that keys are matched by query, and $V$ represents value, which is useful information learned, ${d_k}$ is the dimension of $k$.

Segmenter \cite{segmenter} can better capture the global context information of images. In this paper, it is used as a decoder, as shown in Equation (3).
\begin{equation}
Masks({z_{mask}},c) = softmax(\frac{{{z_{mask}}{c^T}}}{{\sqrt D }})\end{equation}

 In Equation (3), the patch embedding $z_{mask} \in R^{N \times D }$ and class embeddings $c \in {R^{N \times D }}$. The mask obtained is the downsampling graph of N categories, and then upsampling is carried out to restore the size of the original input picture.

PAR \cite{par} module introduced in this paper is used to further refine the fused pseudo label, so as to better align with RGB images and accelerate the convergence speed. PAR is built by affinity pair. Affinity kernel is obtained by adding ${k_{rgb}}$ and ${k_{pos}}$, as shown in Equation (4).

\begin{equation}
{k^{ij,kl}} = \frac{{\exp (k_{rgb}^{ij,kl})}}{{\sum {_{(x,y)}\exp (k_{rgb}^{ij,xy})} }} + {\omega _3}\frac{{\exp (k_{pos}^{ij,kl})}}{{\sum {_{(x,y)}\exp (k_{pos}^{ij,xy})} }}\end{equation}
In Equation (4), $ij$ and $kl$ represent adjacent pixels and adjacent positions. In addition, many nearest neighbors with different dilated rates are defined to further refine the pseudo label feature information.

\subsection{End-to-End Architecture}
\label{ssec:subhead}
The proposed DMSA architecture is shown in Figure 2. ASPP of dynamic dilated rate can refine the quality of attention map better and make the prediction effect of CAM more accurate. The output of CAM is shown in Equation (5).
\begin{equation}{L^c} = ReLU(\sum\limits_k {{\alpha _k}^c{A^k}} )\end{equation}
In Equation (5), $A$ represents the output feature of the last convolution, $k$ represents the $k$-th channel of the feature $A$, and ${\alpha _k}^c$ represents the weight value of ${A^k}$.

The model DMSA introduces four loss functions to better guide the training of student network. In addition to the most common cross-entropy loss, it also introduces segmentation loss, uncertainty loss and diversity loss. The total loss of the final model is shown in Equation (6).

\begin{equation}{{\cal L}_{all}}{\rm{ = }}{\lambda _1}{{\cal L}_{seg}} + {\lambda _2}{{\cal L}_{CE}} + {\lambda _3}{{\cal L}_{un}} + {\lambda _4}{{\cal L}_{cls}}\end{equation}

In Equation(6), ${{\cal L}_{CE}}$ stands for cross-entropy loss. It can better measure the difference between model learning and the real distribution. 
${{\cal L}_{seg}}$ represents segmentation loss and it is used to improve the quality of pseudo labels. 
${{\cal L}_{un}}$ represents uncertainty loss and it is used to maximize the difference between the maximum value and the submaximum value. 
${{\cal L}_{cls}}$ represents classification loss and it is used to alleviate model collapse. $\lambda$ is the weight factor of different loss functions between corresponding subscripts. It is used to balance the loss.

\section{Experiments}
\label{sec:typestyle}
\subsection{Experimental Setup}
\label{ssec:subhead}
This paper conducts unsupervised semantic segmentation experiments on 6 NVIDIA RTX3090 GPUs to verify the performance of the proposed model.

Different from the original dilated convolution with different dilated rates, this paper adopts a dynamic dilated rate strategy and defines a dilated strategy table. When k is equal to 0, the dilated rate in this paper is the same as the dilated rate in the original DeepLabV3+, and the other rounds are different combinations of small dilated rates, as shown in Table 1.

\begin{table}[h]\centering
\caption{Different dilated combinations performed by ASPP. \_k means that epoch modulus 10 is k.}
\begin{tabular}{|c|c|}
\hline
Epoch           & Dilated rate    \\ \hline
\_1,\_3,\_5,\_7 & {[}1,1,2,3{]}   \\ \hline
\_2,\_4,\_6     & {[}1,1,3,5{]}   \\ \hline
\_8,\_9         & {[}1,3,6,9{]}   \\ \hline
\_0             & {[}1,6,12,18{]} \\ \hline
\end{tabular}
\end{table}
IIC \cite{iic}, PiCIE \cite{picie}, MaskContrast \cite{maskcontrast}, TransFGU \cite{TransFGU} are compared with DMSA method in this paper. IIC introduces a concept of maximum mutual information and two inputs to enable the model to identify different objects. PiCIE trains the network by clustering pixel-level feature vectors, so that PiCIE does not distinguish foreground and background information during training. MaskContrast uses supervised saliency to get the masks prior to pixel embedding. In this paper, the method borrowed from TransFGU, uses a self-supervised semantic segmentation architecture based on ViT.

\subsection{Main Results}
\label{ssec:subhead}

The training results of this paper on MS COCO 2017 \cite{mscoco} validation set are shown in Table 2. 

\begin{table}[ht]\centering
\caption{Qualitative comparison between different models on the MS COCO 2017 dataset}
\begin{tabular}{cc|cc}
\hline

Dataset                           & Method       & MIoU           & Acc            \\ \hline
\multirow{4}{*}{COCO-Stuff 27}    & IIC          & 2.36           & 21.02          \\
                                  & PiCIE        & 11.88          & 37.20          \\
                                  & TransFGU     & 16.19          & 44.52          \\
                                  & DMSA         & \textbf{17.82} & \textbf{46.43} \\ \hline
                                  & MaskContrast & 3.73           & 8.81           \\
COCO 80                           & TransFGU     & 12.69          & 64.31          \\
                                  & DMSA         & \textbf{14.69} & \textbf{69.70} \\ \hline
\end{tabular}
\end{table}

As can be seen from Table 2, compared to TransFGU on COCO-Stuff 27, the training effect of the proposed model is significantly improved, where Miou is improved by 1.78 and Acc is improved by 1.09. In particular, the effect is more obvious on COCO 80, where Miou and Acc indicators are improved by 2.0 and 5.39. While improving the effect, the convergence speed of the model in this paper is also greatly improved. Compared with the original 74 epochs, it only needs 6 epochs to converge now.

The visualization effect on COCO-Stuff 27 is shown in Figure 3, it can be seen that DMSA is better than PiCIE and TransFGU in segmenting rule targets such as people, animals and cars. The image in the second column of the first row shows that Picie made mistakes in the segmentation of people. The image in the third column of the second row and the image in the third column of the third row show that the segmentation of bear and car by TransFGU method will result in unclear objects. These problems can be avoided in DMSA.

\begin{figure}[h]
\begin{minipage}[b]{1.0\linewidth}
  \centering
  \centerline{\includegraphics[width=6.5 cm]{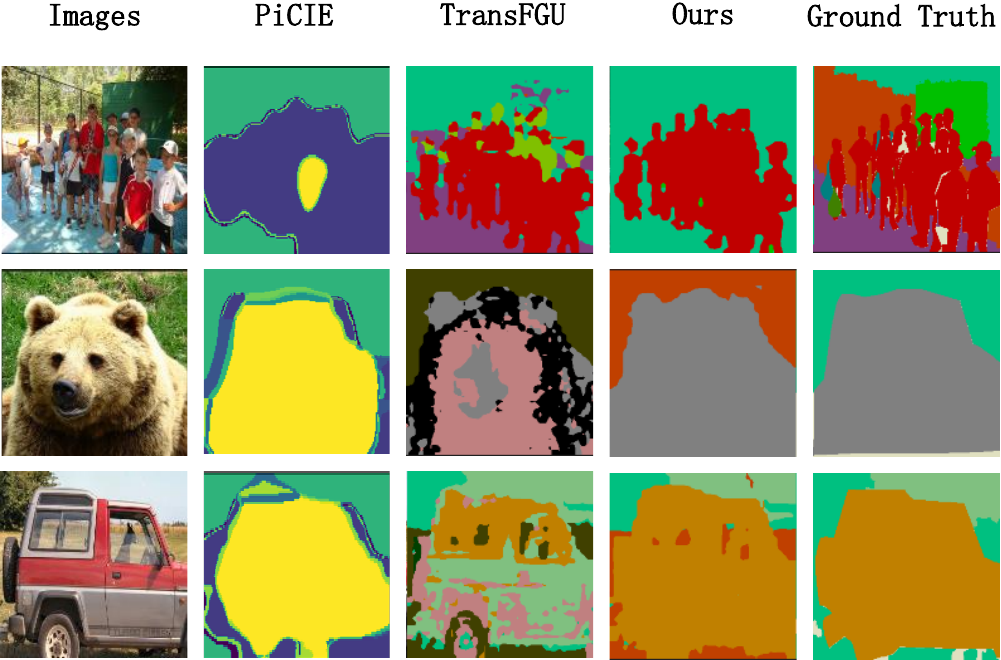}}
\end{minipage}
\caption{Comparison of visualization results between different methods on COCO-Stuff 27.}
\label{fig:res}
\end{figure}
The performance comparison of the proposed model and other models on the Pascal VOC\cite{pascalvoc} validation set is shown in Table 3. 
\begin{table}[ht]\centering
\caption{Comparison between DMSA and different methods on Pascal VOC dataset}
\begin{tabular}{cc|cc}
\hline
Dataset                  & Method       & MIoU           & Acc            \\ \hline
                         & MaskContrast & 35.00          & 79.84          \\
Pascal VOC 2012           & TransFGU     & 37.15          & 83.59          \\
                         & DMSA         & \textbf{37.20} & \textbf{83.61} \\ \hline
Pascal VOC 2012           & TransFGU     & 33.14          & 79.70          \\
Augmented                & DMSA         & \textbf{38.13} & \textbf{83.10} \\ \hline
\end{tabular}
\end{table}

As can be seen from Table 3, the performance of the proposed model is superior to TransFGU on the Pascal VOC 2012 dataset. The training speed of the proposed model is improved. Compared to TransFGU in 60-80 epochs, the proposed method only needs 16 epochs to obtain the similar training effect of TransFGU. On the Pascal VOC 2012 Augmented dataset, MIoU and Acc are improved by 4.9 and 3.4, respectively. In addition, compared to TransFGU converges in the 77th epoch, this method only converges to the maximum value in the 28th epoch. A visual comparison on Pascal VOC 2012 Augmented \cite{pascalvoc,sbd} is shown in Figure 4.

\begin{figure}[h]
\begin{minipage}[b]{1.0\linewidth}
  \centering
  \centerline{\includegraphics[width=6.5 cm]{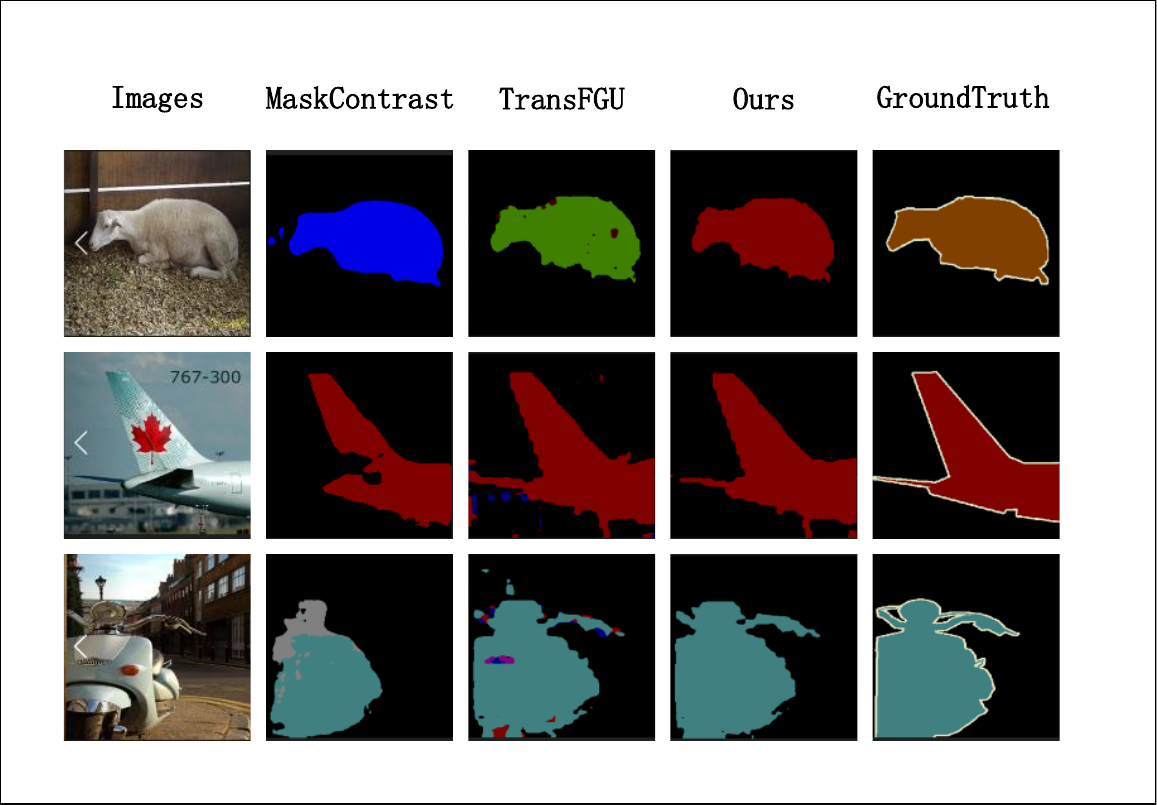}}
\end{minipage}
\caption{Comparison of segmentation effects of different models on Pascal VOC 2012 Augmented dataset.}
\label{fig:res}
\end{figure}

As can be seen from Figure 4 that the proposed method has better effect on the edge refinement problem. It can be seen that MaskContrast may have the problem about incomplete prediction objects, while TransFGU will have the problem that the prediction area exceeds the target area during prediction. These problems can be better avoided in DMSA.

\subsection{Ablation Studies}
\label{ssec:subhead}

In order to verify the effectiveness of each module in this paper, this paper uses ViT-S/16 and ViT-S/8 as the encoder backbone on Pascal VOC 2012 and COCO 80, and ${{\cal L}_{{\rm{seg}}}}$, ASPP and PAR modules are introduced, as shown in Table 4.

\begin{table}[h]\centering\tabcolsep=0.1cm
\caption{Qualitative comparison of different backbone architectures and modules used on the validation set.}
\begin{tabular}{c|c|ccc|cc}
\hline
Dataset   & Method   & ${{\cal L}_{all}}$        & ASPP                      & PAR                       & MIoU           & Acc            \\ \hline
\multirow{6}{*}{COCO 80}   & ViT-S/16 &                           &                           &                           & 11.87          & 59.34          \\ \cline{2-7} 
          &          &                           &                           &                           & 12.57          & 64.06          \\
          &          & \Checkmark &                           &                           & 13.46          & 68.80          \\
          & ViT-S/8  &                           & \Checkmark &                           & 13.17          & 64.85          \\
          &          &                           &                           & \Checkmark & 12.58          & 68.50          \\
          &          & \Checkmark & \Checkmark & \Checkmark & \textbf{14.69} & \textbf{69.70} \\ \hline
          & ViT-S/16 &                          &                           &                           & 30.47          & 69.37          \\ \cline{2-7} 
          &          &                           &                           &                           & 35.08          & 81.20          \\
Pascal VOC &          & \Checkmark &                           &                           & 35.51          & 81.80          \\
2012      & ViT-S/8  &                           & \Checkmark &                           & 37.11          & 82.87          \\
          &          &                           &                           & \Checkmark & 36.16          & 83.15          \\
          &          & \Checkmark & \Checkmark & \Checkmark & \textbf{37.20} & \textbf{83.60} \\ \hline
\end{tabular}
\end{table}

As can be seen from Table 4, ViT-S/8 is better than ViT-S/16 in the backbone of DMSA. The effect after the introduction of ${{\cal L}_{all}}$, ASPP and PAR is improved compared with that before the introduction. The best effect can be achieved when all modules are introduced.

\section{Conclusion}
\label{sec:print}
This paper designs a dynamic and adaptive Transformer unsupervised semantic framework. The ASPP module is introduced into the attention part of the Transformer encoder and a dynamic inflation rate strategy is designed, which makes use of different inflation rate combinations among epochs to better representation learning. In addition, the RAP module is introduced, which is used to refine the fused labels, ensure better alignment with the image information, and greatly improve the convergence speed while improving the training effect. On this basis, the total loss composed of four loss functions can better guide the training process. The experiment shows that the end-to-end unsupervised semantic segmentation framework designed in this paper is more effective on the saliency dataset that is easy to distinguish target objects.

\bibliographystyle{IEEEbib}
\bibliography{strings,refs}

\end{document}